\documentclass[pdflatex,sn-apa]{sn-jnl}

\usepackage{graphicx}%
\usepackage{multirow}%
\usepackage{amsmath,amssymb,amsfonts}%
\usepackage{amsthm}%
\usepackage{mathrsfs}%
\usepackage[title]{appendix}%
\usepackage{xcolor}%
\usepackage{textcomp}%
\usepackage{manyfoot}%
\usepackage{booktabs}%
\usepackage{algorithm}%
\usepackage{algorithmicx}%
\usepackage{algpseudocode}%
\usepackage{listings}%
\usepackage[most]{tcolorbox}
\usepackage{cprotect}
\usepackage{verbdef}
\usepackage{enumitem}
\usepackage{fullpage}

\theoremstyle{thmstyleone}%
\theoremstyle{thmstyletwo}%

\theoremstyle{thmstylethree}%

\begin{document}

\title[Article Title]{
AI generates well-liked but templatic empathic responses
}


\author*[1]{\fnm{Emma} \spfx{S.} \sur{Gueorguieva}}\email{emmagueorguieva@utexas.edu}

\author[2]{\fnm{Hongli} \sur{Zhan}}\email{honglizhan@utexas.edu}

\author[3]{\fnm{Jina} \sur{Suh}}\email{jinasuh@cs.washington.edu}

\author[4]{\fnm{Javier} \sur{Hernandez}}\email{javierh@microsoft.com}

\author[5]{\fnm{Tatiana} \sur{Lau}}\email{tatiana.lau@tri.global}

\author[2]{\spfx{Junyi} \fnm{Jessy} \sur{Li}}\email{jessy@austin.utexas.edu}

\author*[1]{\fnm{Desmond} \spfx{C.} \sur{Ong}}\email{desmond.c.ong@gmail.com}

\affil[1]{\orgdiv{Department of Psychology}, \orgname{The University of Texas at Austin}
}

\affil[2]{\orgdiv{Department of Linguistics}, \orgname{The University of Texas at Austin}}

\affil[3]{\orgdiv{Department of Computer Science and Engineering}, \orgname{The University of Washington}}

\affil[4]{\orgname{Microsoft Research}
}

\affil[5]{\orgname{Toyota Research Institute}
}

\abstract{

Recent research shows that greater numbers of people are turning to Large Language Models (LLMs) for emotional support, and that people rate LLM responses as more empathic than human-written responses. We suggest a reason for this success: LLMs have learned and consistently deploy a well-liked template for expressing empathy. We develop a taxonomy of 10 empathic language ``tactics" that include validating someone's feelings and paraphrasing, and apply this taxonomy to characterize the language that people and LLMs produce when writing empathic responses. Across a set of 2 studies comparing a total of $n$ = 3,265 AI-generated (by six models) and $n$ = 1,290 human-written responses, we find that LLM responses are highly formulaic at a discourse functional level. 
We discovered a template—a structured sequence of tactics—that matches between 83--90\% of LLM responses (and 60-83\% in a held-out sample), and when those are matched, covers 81--92\% of the response. By contrast, human-written responses are more diverse. We end with a discussion of implications for the future of AI-generated empathy.

}


\maketitle

\section{Introduction}\label{sec:Intro}


In times of distress, we often turn to others for empathy \citep{decety2014social, brown2021emotional}, and receiving such social support is crucial for well-being \citep{berkman2000social}. Recent growth in the use of Large Language Models (LLMs) has been accompanied by a new trend: people are increasingly seeking empathy and support from LLMs, especially for companionship and therapy \citep{inzlicht2024praise, mcbain2025use, stade2025current}. Recent research has found that people reliably rate LLM-generated responses to support-seeking posts as more empathic, compassionate, and as helping people feel more heard than human-written responses across a wide range of contexts \citep{lee2024large, yin2024ai, ong2026ai}---even when such responses are written by crisis responders \citep{ovsyannikova2025third}. Other research has found that chatting with LLMs may be associated with improved mental health \citep{siddals2024happened}, reduced loneliness \citep{de2024ai}, and even reduced suicidal ideations for some users \citep{maples2024loneliness}. A recent clinical trial with a generative AI-powered therapy chatbot has also suggested that it can reduce symptoms of depression, anxiety, and eating disorders \citep{heinz2025randomized}.

Given these emerging findings, it is surprising that not many studies have gone beyond subjective ratings to investigate what is actually in the language of these LLM-generated empathic messages. Studies have noted that LLMs produce longer messages than people \citep{zhang2024verbosity} or choose longer words in favor of more abbreviated words (e.g., “mathematics” over “math”) \citep{cai2024large}, which may be interpreted by empathy recipients as a measure of empathic effort or concern \citep{cameron2019empathy, zaki2014empathy}. Only one study \citep{lee2024large} applied language analyses to identify n-grams that are most predictive of perceived empathy. A focus on the language—the concrete “behaviors” that are exchanged between LLMs and their human users—is important and can complement studies on the psychological perceptions of AI-generated empathy \citep{rubin2025value, wenger2026people}.

One reason why a focus on language is important in studying perceived empathy is that there exists a gap between psychological theories and what is required for AI models. Existing psychological theories of empathy focus primarily on empathizer capacities like the ability to share another’s feelings (experience sharing; \cite{decety2008emotion, shamay2009two}), the ability to take another’s perspective (perspective taking; \cite{galinsky2008pays, lamm2007neural}), or the desire to help someone in distress (empathic concern; \cite{batson2011altruism, zaki2014empathy}). But scholars have pointed out that AI has no emotions and so, according to these definitions, cannot truly empathize \citep{perry2023ai, montemayor2022principle}. Yet, this does not jive with growing evidence from psychology \citep{ong2026ai} and medicine \citep{howcroft2025ai} that people consistently rate AI-generated responses as more empathic than human-written responses \citep{lee2024large, ovsyannikova2025third, yin2024ai}. The evidence for this main effect is overwhelming: two recent reviews found that 19 of 19 effect sizes \citep{ong2026ai} and 13 of 15 studies \citep{howcroft2025ai} found evidence in favor of AI. But psychological research also finds a second main effect: controlling for who (or what) generated the response, when people are \emph{told} that a response had been generated by an AI, they perceive it as being less empathic \citep{ovsyannikova2025third, rubin2025value, yin2024ai}. This second main effect is purely psychological—perhaps relating to people’s judgments that AI lacks emotions or other emotional capacities, and so AI-generated empathy is less credible than human empathy \citep{jackson2023exposure, oldemburgo2025moralization}. But the first main effect suggests a testable hypothesis about objectively measurable behavior: LLMs produce language that is of higher empathic quality than those produced by people, and we should be able to identify patterns of language that support the subjective quality that people perceive in LLM-generated responses. This motivated the present focus on the language behaviors that support human perceptions of empathy.

Other studies in non-empathy domains have found that LLM responses, especially for open-ended queries, lack content diversity \citep{jiang2025artificial}: They seem to generate repetitive phrases or overuse certain words within messages \citep{reinhart2025llms}, and produce responses that follow syntactic \citep{shaib2024detection} and discourse \citep{namuduri2025qudsim} templates. The resulting language, therefore, can be regarded as “slop” \citep{shaib2025measuring}.
The lack of diversity in AI-generated content can also negatively impact human creative ideation \citep{anderson2024homogenization,doshi2024generative,gerlich2025ai}, and consistent exposure to low quality, repetitive content may put users at risk of negative emotional and cognitive consequences \citep{nolan_kimball_2026}.
Based on these findings, we expect to see similar findings when LLMs are prompted to produce emotionally supportive messages: they produce templated responses that are much more homogeneous compared to human-generated messages.

In this paper, we first introduce a taxonomy of ten empathic response “tactics” that are explicitly designed to be identifiable in text—for instance, phrases that validate the other’s person feelings (validation); disclose a similar experience that the empathizer had gone through (self-disclosure); or offering advice. The focus on identifying objective behaviors is in contrast to previous studies that relied only on subjective reports of perceived empathy of a response \citep{ovsyannikova2025third, welivita2024large}, or asking human raters to read a response and rate if the response writer used a specific strategy (e.g., Study 2 from \cite{yin2024ai}). Our fine-grained, text-level tactics taxonomy also differs from prior work in computationally modeling empathy, such as \cite{sharma2020computational}’s broad categories of “Emotional Reactions”, “Explorations”, and “Interpretations”; \cite{suh2026sense}’s seven category taxonomy of perceptions of AI empathy; and \cite{iyer2026heart}'s five categories for benchmarking AI-generated empathy. These, and other previous frameworks \citep{hu2024aptness, lee2024comparative}, are all broader and more abstract characterizations of empathy that capture various aspects of empathic behavior—for instance, an empathic agent shows perspective taking, or contextual understanding \citep{suh2026sense, iyer2026heart}. Our taxonomy complements these previous frameworks by identifying specific ways that these constructs are expressed in language. Perhaps the closest is previous computational work that have tried to define and build classifiers for specific support strategies \citep{liu2021towards} or “empathic response intents” \citep{welivita2020taxonomy}, although ours is based more on broader psychological theories of empathy and social support than just a single paper (see Methods for our approach).

Armed with our taxonomy, we proceeded to analyze the prevalence of these tactics in LLM-generated and human-written responses. In Study 1, we apply our taxonomy to a sample of 290 human-written and 303 LLM-generated responses from 3 models, and we had human raters manually annotate the presence of tactics at a sub-sentence level (i.e., a sentence could have multiple tactics; raters identified the specific phrases that corresponded to each tactic). In Study 2, we then prompted an LLM classifier to annotate the presence of tactics, which allowed us to analyze a larger dataset of 1,000 human-written and 2,962 LLM-generated responses from 3 newer models. Across both studies, we show significant differences between LLMs and humans---but surprising consistency across LLMs---in the way that writers use empathic tactics. We identify “templates” inspired by regular expressions that capture a significant majority of LLM, but not human, responses. 
Finally, we conclude with a discussion of how our results relate to recent work on understanding of LLM-generated language, and to our understanding of empathic expressions more broadly.

\section{Results}\label{sec:Results}

\begin{table}[!ht]
\centering
\begin{tabular}{p{0.09\linewidth}p{0.13\linewidth}p{0.35\linewidth}p{0.32\linewidth}}
\hline
\textbf{Empathy Facet} & \textbf{Tactic}   & \textbf{Description}    & \textbf{Examples}  \\\hline
\multirow{3}{=}{Experience Sharing} &
\hangindent=0.5em Emotional E\textbf{x}pression      & \hangindent=0.5em Communicating the empathy-giver's feelings, reactions, or thoughts   &  \hangindent=0.5em \emph{I'm so sorry to hear that} $|$ \emph{Wow, what a beautiful story} \\
& \textbf{E}mpowerment       & \hangindent=0.5em Positive, uplifting statements about the empathy-seeker's character and capabilities   &  \hangindent=1em \emph{You are going to get through this.} \\
& \textbf{V}alidation        & \hangindent=0.5em Reassures, normalizes, or validates an empathy-seeker's feelings   &  \hangindent=0.5em \emph{Everyone has feelings like this.} $|$ \emph{You're not overreacting.} \\
\hline
\multirow{4}{=}{Perspective Taking} &
\textbf{I}nformation       & \hangindent=0.5em Offering facts or resources (e.g., links) &  \hangindent=0.5em \emph{Flying is the safest form of travel.} \\
& \textbf{P}araphrasing      & \hangindent=0.5em Restating something the empathy seeker said to demonstrate understanding of their situation, feelings, or experiences   & \hangindent=0.5em \emph{I'm hearing that you feel overwhelmed} \\
& \textbf{R}eappraisal       & \hangindent=0.5em Helping to engage in cognitive reappraisal (changing a belief)   & \hangindent=0.5em \emph{that was out of your control} \\
& \hangindent=0.5em \textbf{S}elf-Disclosure   & \hangindent=0.5em Sharing personal information or similar past experiences or feelings   & \hangindent=0.5em \emph{I've had that happen to me before too} \\
\hline
\multirow{3}{=}{Empathic Concern} &
\textbf{A}dvice            & \hangindent=1em Providing ideas for solutions or coping strategies   & \hangindent=1em \emph{If I were you I would see a therapist} $|$ \emph{get some ice cream!} $|$ \emph{Definitely talk to your boss} \\
& Assis\textbf{t}ance        & \hangindent=0.5em Offering some aid to the empathy-seeker   &  \hangindent=0.5em \emph{I'm here for you if you want to talk} $|$ \emph{Can I do anything to help?} \\
& \textbf{Q}uestioning       & \hangindent=0.5em Asking questions to improve understanding of the empathy-seeker's feelings, experiences, or situations.   &  \hangindent=0.5em \emph{How are you feeling?} $|$ \emph{What do you think about [x]?} \\
\hline
\end{tabular}
\caption{Taxonomy of tactics, along with description and examples. Bolded letters in tactics indicate abbreviations used in the paper. Note: The full codebook for identifying these tactics in text is given in the Supplemental Material.}
\label{tab:taxonomy}
\end{table}


\subsection{LLM-generated text are perceived as empathic}

Study 1 compared three LLMs (GPT-4 Turbo, Llama3.1-70b, and Qwen2.5), with humans with a psychology background, in responses to 101 support-seeking Reddit posts, and with fully human-annotated tactics (see Methods for annotation details). First, we examined human perceptions of empathy from these responses: an independent set of human raters perceived find that on average, human-written responses ($M_{human}$ = 3.71, 95\% CI = [3.63, 3.79]) were perceived to be less empathic than responses written by LLMs ($M_{GPT-4 Turbo}$ = 4.18 [4.07, 4.29], Cohen’s $M_{dgpt > human}$ = 0.39, $t_{tgpt > human}$ = 8.72,  $p <$ .001; $M_{llama3.1}$ = 4.14, [4.03, 4.26], $d_{llama > human}$ = 0.37, $t_{llama > human}$ = 7.87,  $p <$ .001; $M_{Qwen2.5}$ = 4.20 [4.09, 4.31], $d_{qwen > human}$ = 0.45, $t_{qwen > human}$ = 8.98,  $p <$ .001). These results replicate, and are surprisingly numerically consistent with, a well-known finding that people rate text generated by LLMs to be more empathic than text written by humans—e.g., see \cite{ong2026ai} who estimated a meta-analytic Hedges’ $g$ of 0.54 from 19 effect sizes (see also \cite{howcroft2025ai}). While not necessarily novel, these perceived empathy ratings serve as background to contextualize our next set of language findings.

Replicating other findings on this topic, we found that, on average, human-written responses were perceived as less empathic than responses written by any of our LLM writers ($M_{human}$ = 3.71, 95\% CI = [3.63, 3.79]; $M_{GPT-4 Turbo}$ = 4.18, [4.07, 4.29], \emph{t} = 8.72,  $p < .001$; $M_{Llama3.1}$ = 4.14, [4.03, 4.26], \emph{t} = 7.87,  $p < .001$; $M_{Qwen2.5}$ = 4.20, [4.09, 4.31], \emph{t} = 8.98,  $p < .001$). We found the same pattern for appropriateness ratings.

\subsection{LLMs use a relatively homogeneous set of tactics.}

\begin{table}[!ht]
    \centering
    \begin{tabular}{lrrr}
    \hline
    & Word count & Total number of tactics & Unique tactics \\
    Study 1 Humans (Upworkers)    & 234 (141) & 12.2 (6.48) & 5.49 (1.51) \\
    \texttt{GPT-4 Turbo}  & 234 (114) & 15.9 (7.64) & 5.50 (1.07) \\
    \texttt{Llama3.1-70b} & 147 (33.5) & 10.5 (3.55) & 5.13 (1.21) \\
    \texttt{Qwen2.5}      & 166 (36.0) & 10.9 (3.21) & 
    5.25 (0.97) \\
    \hline
    Study 2 Humans (Redditors)    & 183 (98.8) & 7.20 (4.98) & 4.29 (1.89) \\
    \texttt{GPT-4o}       & 179 (8.0)  & 11.3 (2.62) & 5.16 (0.99) \\
    \texttt{Llama3.3-70b-instruct} & 168 (25.6) & 9.25 (2.30) & 4.63 (1.06) \\
    \texttt{Qwen3-32b}        & 120 (16.2) & 12.0 (3.03) & 5.54 (0.99) \\
    \hline
    \end{tabular}
    \caption{Descriptive statistics of the empathic responses across both studies. Numbers represent mean, with standard deviations in parentheses. Unique tactics reflects the mean number of unique tactics used per response. Note that human writers were not given a target word limit, and LLMs were given a word limit of 150 words (see Methods)}
    \label{tab:descriptives}
\end{table}

\begin{figure}[tbh]
    \centering
    \includegraphics[width=\textwidth]{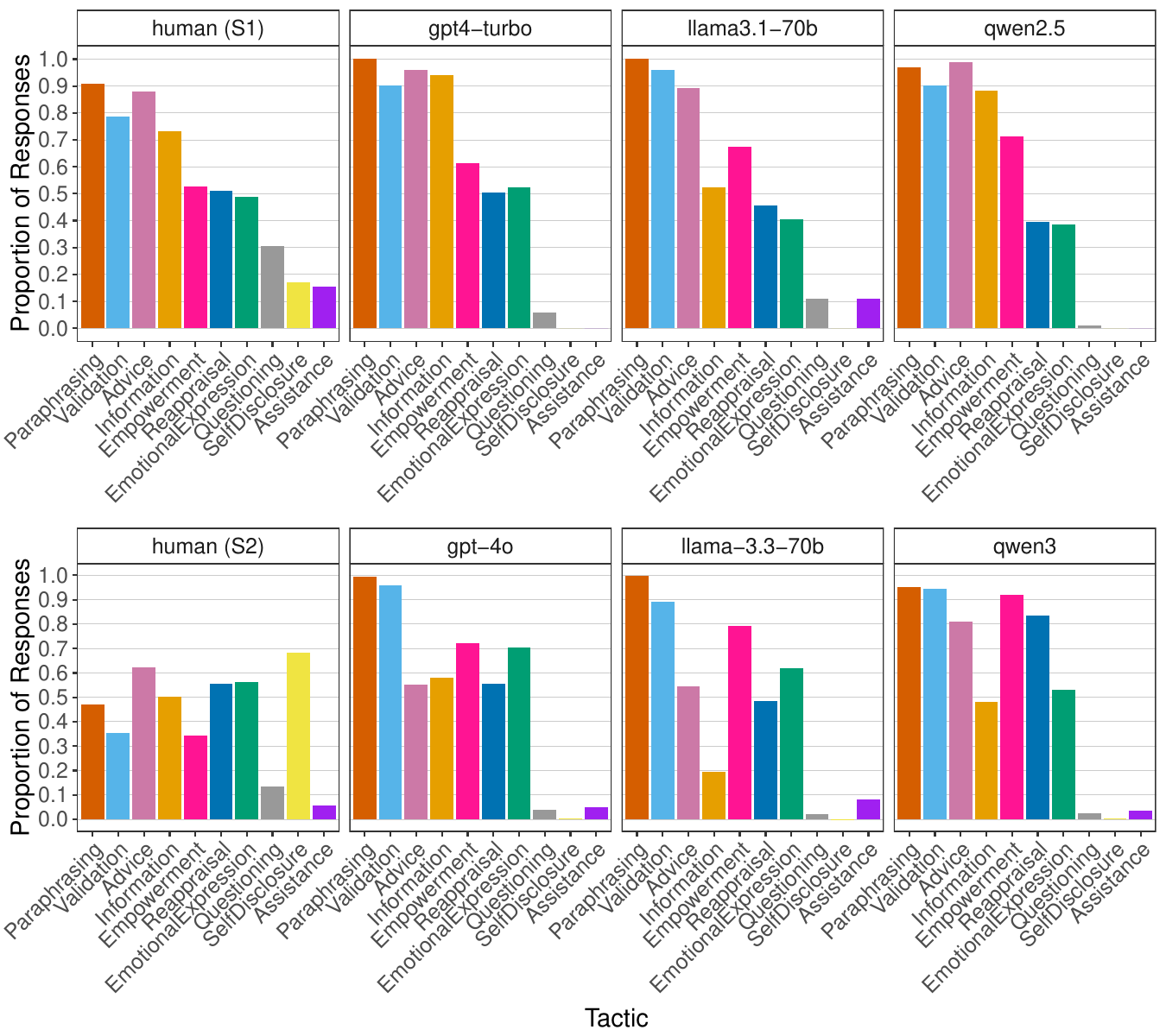}
    \caption{Distribution of unique empathic tactic usage across writers in Study 1 (top) and Study 2 (bottom). Here, we consider unique tactics: that is, how many responses contained at least one of each tactic type. 
    }
    \label{fig:tactic_prevelance}
\end{figure}

Next, we analyzed the prevalence of empathic tactics across all responses. On average, our human writers wrote messages of about 234 words in length, while LLMs produced between 147 and 234 words (see Table \ref{tab:descriptives}). The differences in word counts did translate to corresponding differences in total number of tactics; but on average all writers used a similar number of unique tactics per response. 

Despite using similar numbers of tactics, the distribution of tactics across humans and LLMs is different. LLMs tend to use a much smaller and less diverse set of tactics (see Fig. \ref{fig:tactic_prevelance}). \texttt{GPT-4 Turbo} uses Paraphrasing (100\%), Advice (96.0\%), Validation (90.1\%), and Information (94.1\%), in over 90\% of responses. These prevalences were also reflected in \texttt{Llama3.1}, which used Paraphrasing (100\%), Advice (89.1\%), and Validation (96.0\%) in almost all responses, and in \texttt{Qwen2.5}, which similarly used Paraphrasing (97.0\%), Advice (99.0\%), Validation (90.1\%), and Information (88.1\%) in almost all responses. While human writers also frequently used Paraphrasing (91.0\%) and giving Advice (88.6\%), they tend to use a much broader range of tactics overall. This pattern is also seen at the low-prevalence end: there are certain tactics that people use that almost never show up in LLM responses, such as Questioning (M$_\texttt{gpt-4 turbo}$ = 5.9\%, M$_\texttt{llama3.1}$ = 10.9\%, M$_\texttt{qwen2.5}$ = 0.9\%) and Assistance (M$_\texttt{gpt-4 turbo}$ = 0\%, M$_\texttt{llama3.1}$ = 10.9\%, M$_\texttt{qwen2.5}$ = 0\%). Models also did not offer any Self-Disclosure (0\% across all models; compared to 17.2\% for humans); although this may be understandable as a model design choice—people might feel that it is strange or that it is inherently deceptive if an LLM reports any personal experiences—sharing personal experiences is a crucial part of human empathy that scholars have pointed out that AI lacks \citep{perry2023ai}. Interestingly, the rank-order correlation of tactic prevalence produced by the models with those produced by humans is high ($r_{human-gpt}$ = .98, $r_{human-llama}$ = .94, $r_{human-qwen}$ = .97)---if people tend to use more of a certain tactic, models do too, which is not surprising as models are trained on human text, but the diversity is far less than humans.

In Study 2, with a different sample of LLMs and humans, we find very similar results for LLMs, and very different results for humans. For LLMs, we used updated versions of the same model families, specifically, \texttt{GPT-4o}, \texttt{Llama3.3-70b-instruct}, and \texttt{Qwen3-32b}, to respond to a larger set of 1000 Reddit posts. Overall, models still overwhelming used Paraphrasing (M$_\texttt{gpt-4o}$ = 99.3\%, M$_\texttt{llama3.3}$ = 99.8\%, M$_\texttt{qwen3}$ = 95.2\%) and Validation (M$_\texttt{gpt-4o}$ = 96.0\%, M$_\texttt{llama3.3}$ = 89.1\%, M$_\texttt{qwen3}$ = 94.6\%). Interestingly, compared to Study 1, there was a drop in Advice (M$_\texttt{gpt-4o}$ = 55.2\%, M$_\texttt{llama3.3}$ = 54.6\%, M$_\texttt{qwen3}$ = 81.0\%) and Information (M$_\texttt{gpt-4o}$ = 58.1\%, M$_\texttt{llama3.3}$ = 19.6\%, M$_\texttt{qwen3}$ = 48.0\%). The overall tactic prevalences within model families were highly correlated ($r_{gpt4: turbo, 4o}$ = .88, $r_{llama:3.1, 3.3}$ = .89, $r_{qwen:2.5, 3}$ = .85).

For human writers, we wanted a more “naturalistic” sample of responses, and so we collected the top-rated response that was at least 100 words long, to those same Reddit posts that LLMs responded to. These Reddit comments were shorter on average than our Upworker responses from Study 1, and used much fewer tactics (Table \ref{tab:descriptives}). But the distribution of human responses was also very different in Study 2, compared to Study 1 (Fig. \ref{fig:tactic_prevelance}). For one, Redditors provided less paraphrasing, advice, and validation overall. They did, however, self-disclose a lot more (68.2\%)---this is understandable, as these are people who are voluntarily responding to strangers on an internet forum, and one motivation to do so is to share what they themselves have gone through as well. Overall, the tactic prevalences across Study 1 and 2 humans are not as correlated ($r_{human:s1,s2}$ = .37) as what we see for the LLMs.


\subsection{LLM responses are highly templated.}

Not only do LLMs use the same, relatively homogeneous set of tactics, they also do so in a relatively fixed manner. Because we annotated the appearance of tactics within the response by identifying parts of sentences that reflected a tactic, we have an ordered sequence of tactic codes for each response. We used the language of \textbf{regular expressions} (in shorthand, “regex”) to capture structured patterns of tactic occurrences. For instance, we observed that LLM-generated empathic responses in our sample tended to start off with text that paraphrased what the empathy-seeker wrote, along with validation of their feelings. But in some responses, this could come in the opposite order (validation then paraphrasing), while in others, this could cycle several times (paraphrasing, then validation, then paraphrasing, then validation). The regular expression \verb|[PV]+| captures any repeating combination of \textbf{P}(araphrasing) and \textbf{V}(alidation) that appears at least once, and would match sequences like: \textbf{V}, \textbf{P}, \textbf{VP}, \textbf{VPV}, \textbf{PVPV}, and so forth (but not the empty string or strings containing other letters). Thus, regular expressions offer a formal way of capturing repeating patterns by treating each tactic as an atomic character that can be matched in text (see Methods for a more detailed description). 

We created “templates” by defining regular expressions that aimed to maximize (1) coverage across responses (i.e., how many responses contained a sequence that was matched by this regular expression) and (2) coverage within responses (i.e., how much of a given response was captured by this regular expression). Intuitively, shorter regular expressions would have higher coverage \textit{across} responses as it is easier to find responses that contain a sequence corresponding to these shorter expressions, but they would have poor coverage \textit{within} responses as they capture a smaller portion of the responses. Conversely, longer regular expressions are harder to match to responses (i.e., lower coverage \textit{across} responses), but if there is a match, would cover a larger portion of the response (higher coverage \textit{within} response). These regular expressions were discovered via a mix of manual inspection of the data guided by a beam search procedure on Study 1 data where we aimed to maximize both across coverage and within coverage.

\verbdef{\patternOne}{^X?[PV]+ [XE]? [AIP]+ }
\verbdef{\patternTwo}{^X?[PV]+ [XE]? [AIP]+ [VXER]+ }
\verbdef{\patternThree}{^X?[PV]+ [XE]? [AIP]+ [VXER]+ [AIP]+ }
\verbdef{\patternFour}{^X?[PV]+ [XE]? [AIP]+ [VXER]+ [AIP]+ [VXER]+ }
\verbdef{\patternFive}{^X?[PV]+ [XE]? [AIP]+ [VXER]+ [AIP]+ [VXER]+ [AIP]+ }

\verbdef{\patternVXER}{[VXER]+}
\verbdef{\patternAIP}{[AIP]+}

\begin{table}[htb!]
    \scalebox{1.0}{
    \centering
    \begin{tabular}{|p{.08\linewidth}|cccc||cccc|}
        \hline

    & \multicolumn{4}{c||}{Study 1} & \multicolumn{4}{c|}{Study 2} \\
    & Human & \texttt{GPT-4 Turbo} & \texttt{Llama3.1} & \texttt{Qwen2.5} & Human & \texttt{GPT-4o} & \texttt{Llama3.3} & \texttt{Qwen3} \\
    Pattern 1 & \multicolumn{8}{p{0.75\linewidth}|}{
    {\color{red} \patternOne } \newline
    Starting with 0 or 1 EmotionalE\textbf{x}pression, then alternating [\textbf{P}araphrasing and \textbf{V}alidation]; 0 or 1 [EmotionalE\textbf{x}pression or \textbf{E}mpowerment]; alternating [\textbf{A}dvice, \textbf{I}nformation, \textbf{P}araphrasing]} \\
    &&&& \multicolumn{1}{c}{} &&&&\\
    Across & 61.0 & 88.1 & 92.1 & 91.1 & 11.7 & 71.9 & 87.9 & 61.9 \\
    Within & 46.7 & 50.1 & 58.3 & 60.1 & 44.8 & 49.8 & 55.1 & 35.6 \\
    \hline

    Pattern 2 & \multicolumn{8}{p{0.75\linewidth}|}{
    \patternOne {\color{red} \patternVXER} \newline
    Pattern (2) = Pattern (1) plus alternating [\textbf{V}alidation, Emotional E\textbf{x}pression, \textbf{E}mpowerment, \textbf{R}eappraisal]
    \newline
    \textbf{Test: Pattern 2 or 1\$}} \\
    &&&& \multicolumn{1}{c}{} &&&&\\
    Across & 52.4 & 87.1 & 90.1 & 91.1 & 8.94 & 71.4 & 87.1 & 61.6 \\
    Within & 56.5 & 53.0 & 65.7 & 69.7 & 61.1 & 61.0 & 69.3 & 52.4 \\
    \hline
    
    Pattern 3 & \multicolumn{8}{p{0.75\linewidth}|}{
    \patternTwo {\color{red} \patternAIP} \newline
    Pattern (3) = Pattern (2) plus alternating [\textbf{A}dvice, \textbf{I}nformation, \textbf{P}araphrasing]
    \newline
    \textbf{Test: Pattern 3 or 2\$ or 1\$}} \\
    &&&& \multicolumn{1}{c}{} &&&&\\
    Across & 48.3 & 86.1 & 88.1 & 91.1 & 7.73 & 70.8 & 86.0 & 61.4 \\
    Within & 67.7 & 74.2 & 79.2 & 81.7 & 73.9 & 72.9 & 77.8 & 61.2 \\
    \hline
    
    Pattern 4 & \multicolumn{8}{p{0.75\linewidth}|}{
    \patternThree {\color{red} \patternVXER} \newline
    Pattern (4) = Pattern (3) plus alternating [\textbf{V}alidation, Emotional E\textbf{x}pression, \textbf{E}mpowerment, \textbf{R}eappraisal]
    \newline
    \textbf{Test: Pattern 4 or 3\$ or 2\$ or 1\$}} \\
    &&&& \multicolumn{1}{c}{} &&&&\\
    Across & 42.1 & 83.2 & 86.1 & 90.1 & 7.03 & 69.6 & 84.5 & 60.8 \\
    Within & 76.7 & 75.9 & 84.4 & 86.4 & 83.7 & 79.1 & 87.8 & 75.4 \\
    \hline
    
    Pattern 5 & \multicolumn{8}{p{0.75\linewidth}|}{
    \patternFour {\color{red} \patternAIP} \newline
    Pattern (5) = Pattern (4) plus alternating [\textbf{A}dvice, \textbf{I}nformation, \textbf{P}araphrasing]
    \newline
    \textbf{Test: Pattern 5 or 4\$ or 3\$ or 2\$ or 1\$}} \\
    &&&& \multicolumn{1}{c}{} &&&&\\
    Across & 40.3 & 83.2 & 85.1 & 90.1 & 6.43 & 68.9 & 82.9 & 59.9 \\
    Within & 81.7 & 84.8 & 89.5 & 92.1 & 90.3 & 87.7 & 91.3 & 81.1 \\

    \hline

    \end{tabular} }
    \cprotect\caption{Regular expressions representing candidate templates and the proportion of responses they matched across writers (Across Coverage) and, for matched responses, how much of the response is matched by the expression (Within Coverage). Left columns: Study 1, Right columns: Study 2. Tactics: X = Emotional Expression; V = Validation; P = Paraphrasing; A = Advice; I = Information; E = Empowerment; R = Reappraisal. Regular expression syntax: \verb|^| start of string, \$ end of string, [] match set, ? 0 or 1 match, + 1 or more matches. See Methods for more details.}
    \label{tab:regex}
\end{table}

\begin{figure}[tbh]
    \centering
    \includegraphics[width=\textwidth]{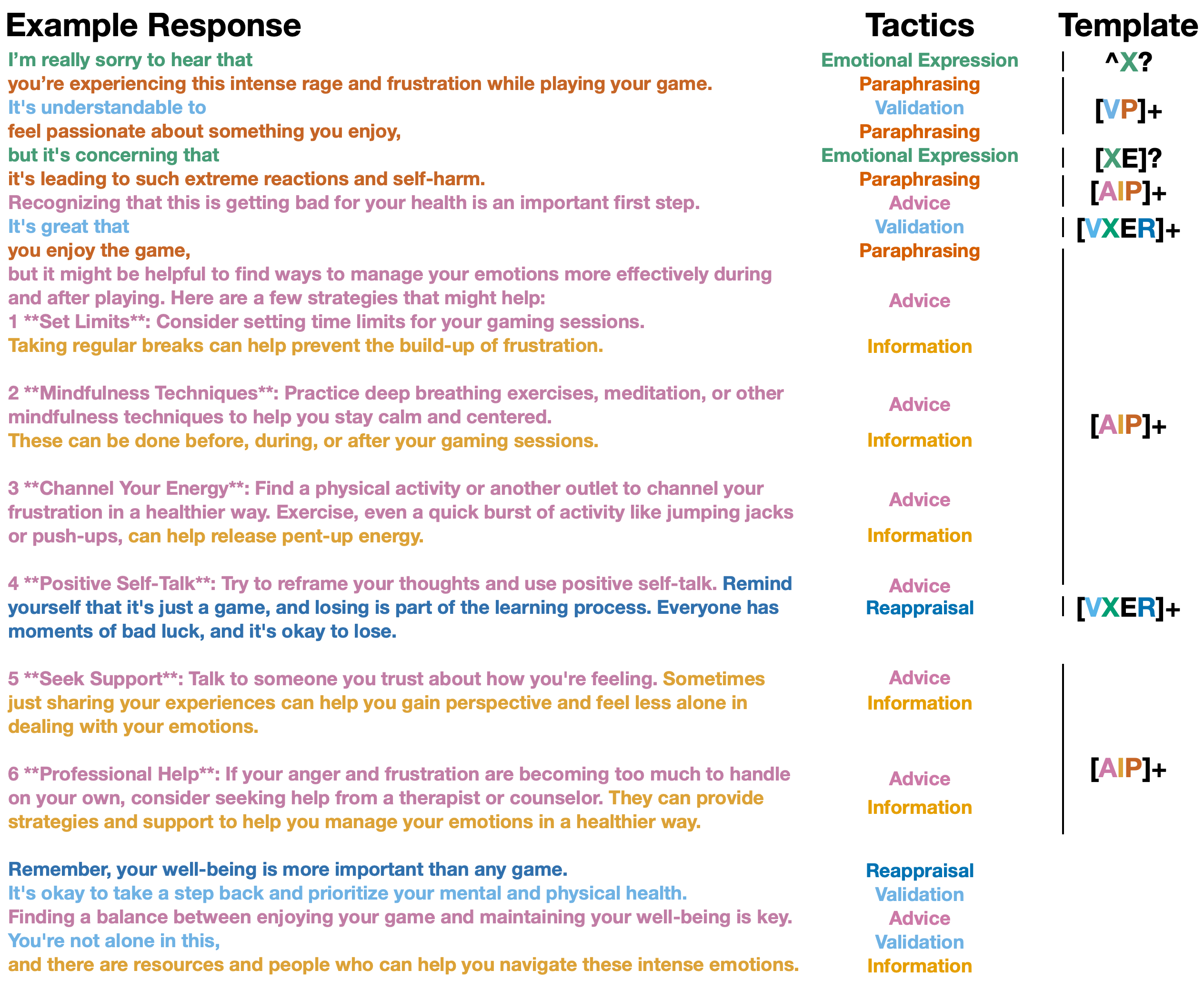}
    \caption{Example response from \texttt{GPT-4 Turbo} with annotated tactics, and how the tactics are matched by Template Pattern 5. Note that we collapse consecutive tactics (e.g., the text including: ``but it might be helpful... here are a few strategies... set limits..." could be tagged as multiple instances of Advice, but we count it as just one instance.). Note that Pattern 5 ends before the response ends, so it only covers 80.8\% of the response (``within coverage"). 
    }
    \label{fig:template}
\end{figure}

LLM responses can be described by a relatively small number of tactic templates (Table \ref{tab:regex}). The vast majority of LLM responses can be described by the following starting pattern (Pattern 1): starting with an optional emotional expression (\verb|X?|), alternating paraphrasing and validation (\verb|[PV]+|), interjecting with an optional emotional expression or a statement of empowerment (\verb|[XE]?|), and then alternating giving advice, information, and paraphrasing (\verb|[AIP]+|). This short template matches 88.0\% of \texttt{GPT-4 Turbo} responses, 92.1\% of \texttt{Llama3.1-7b} responses, and 91.1\% of Qwen2.5 responses (coverage across responses), and of those responses that this pattern matches, this pattern already covers 50.1\% of \texttt{GPT-4 Turbo} responses, 58.3\% of \texttt{Llama3.1} responses, and 60.1\% of \texttt{Qwen2.5} responses (coverage within responses). For humans, this pattern matches 61.0\% of responses, and within those matched responses, covers 46.7\% of the response. 

As we add successively more components to this regular expression template, the coverage across responses decreases, while the coverage within responses increases---but the coverage across LLMs responses do not decrease as quickly as the coverage across human responses. We discovered that the best-matching patterns included adding successive match sets of alternating Validation, Emotional Expression, Empowerment, Reappraisal, (\verb|[VXER]+|), and alternating Advice, Information and Paraphrasing (\verb|[AIP]+|). For each row in Table \ref{tab:regex} where we introduce Pattern $k$, we also include responses that are fully matched by all Patterns $j < k$. Thus, the row for Pattern 2 includes responses matched by Pattern 2 and those fully-matched by Pattern 1 (see Methods for details and justification). By the time we reach Pattern 5, these patterns matched 83.2\% of \texttt{GPT-4 Turbo} responses with a within coverage of 84.8\%; 85.1\% of \texttt{Llama3.1} responses with a within coverage of 89.5\%, and 90.1\% of \texttt{Qwen2.5} responses with a within coverage of 92.1\%. Conversely, these patterns only match 40.3\% of human responses with a within coverage of 81.7\%.

We developed these templates in an iterative process on Study 1 data, which could be overfitted. Hence, we tested the same templates on Study 2 data as a held-out set for replication, and we find striking similarity as well. Pattern 1 matches 71.9\% of \texttt{GPT-4o} responses, 87.9\% of \texttt{Llama3.3} responses, and 61.9\% of \texttt{Qwen3} responses, and when these are matched, the pattern covers 49.8\% of \texttt{GPT-4o} responses, 55.1\% of \texttt{Llama3.3} responses, and 35.6\% of \texttt{Qwen3} responses. By contrast, the pattern only matches a mere 11.7\% of human responses, and when the pattern matches, it covers 44.8\% of the response. As we lengthen the Patterns, the across-coverage decreases slightly, while the within-coverage increases more, such that by the time we get to Pattern 5 (where we test all five nested patterns), we observe across-coverage of 68.9\% for \texttt{GPT-4o}, 82.9\% for \texttt{Llama3.3}, and 59.9\% for \texttt{Qwen3}, but within these matched responses the pattern covers 87.7\% of \texttt{GPT-4o} responses, 91.3\% of \texttt{Llama3.3} responses, and 81.1\% of \texttt{Qwen3} responses. For the human writers in Study 2, these patterns only match 6.4\% of responses, although for those responses the within coverage is high, at 90.3\%.

Thus, the vast majority of LLM responses can be described by this simple pattern, annotated in Fig. \ref{fig:template}: a starting ``introduction" where the LLM mainly offers paraphrasing and validating, with interspersed emotional expressions (“I’m really sorry to hear that” in Fig. \ref{fig:template}) and empowerment (e.g., “you’ve got this!”). This is followed by the main ``body" of the response which alternates between one chunk which provides advice and information, and connecting it back to the user’s experiences (via paraphrasing); and a second occasional chunk with statements offering validation, an emotional expression, empowerment, and/or reappraisal. This template matches 83–90\% of LLM responses in Study 1, and within those responses matches 85–92\% of what they produce. In Study 2, the across-coverage is slightly lower, at 60–83\%, but the within coverage remains high at 81–91\%. These results are also consistent across different model families: \texttt{GPT-4 Turbo} and \texttt{GPT-4o}, \texttt{Llama3.1} and \texttt{Llama3.3}, and \texttt{Qwen2.5} and \texttt{Qwen3}, suggesting that differences in the training data or other training choices did not seem to have an effect on the empathic templates that models learn (Although we note that the biggest difference across Study 1 and 2 is for \texttt{Qwen}. \texttt{Qwen2.5} in Study 1 had the highest matches with our template, but \texttt{Qwen3} had the lowest matches, even from the first pattern, suggesting that there might have been larger differences from \texttt{Qwen2.5} to \texttt{Qwen3}, than for the other model families. \texttt{Llama}, by contrast, looked very similar across versions 3.1 and 3.3). 
Human responses are, by contrast, more diverse, with only 40\% of responses in Study 1, and 6.4\% of responses in Study 2 being matched by this template—for these matched responses, this template matches 82–90\% of the content written by humans. On the one hand, it is not surprising that the within-response coverage can be high: LLMs must have learnt this pattern of expressing empathy from humans, so at least some humans must produce this pattern. Indeed, in Study 1, we specifically recruited humans with a psychology background, some of whom may have been trained on what makes a good response. On the other hand, the low across-response coverage suggests that people are far more diverse with how they express empathy, with their responses not necessarily fitting into a small set of templates. 

%
%
%
%
%
%
%
%
%
%
%
%

\section{Discussion}\label{sec:Discussion}

As greater numbers of people are turning to LLMs for emotional support, and are perceiving the language of LLM-generated responses to be oftentimes more empathic than human-written responses \citep{ong2026ai, ovsyannikova2025third, yin2024ai}, our findings here start to shed light on why. Our results suggest that LLMs have learned and consistently deploy a well-liked template for expressing empathy. Using a taxonomy of 10 empathic language tactics that allows us to characterize language behaviors within a sentence, we show that LLMs—from three different model families and across two generations, for a total of six models—reliably use a limited set of tactics, and in fact use them according to a well-defined template. At the start of this template, LLMs mainly paraphrase what the empathy seeker has said and offer validation, perhaps sprinkled with an emotional expression or a note of empowerment. This demonstrates an understanding of the empathy seeker’s situation and validates their emotions, which builds rapport and trust. Then LLMs go into a loop of offering advice and information to help the empathy seeker through their challenges, and connecting those back to the empathy seeker’s situation (captured by our Paraphrasing tactic), occasionally interspersed with emotional expressions, empowerment, validation or reappraisal. This simple template—described by a small set of regular expressions—is found in between 83-90\% of LLM responses (and 60-83\% in a held-out sample), and in those matched responses, cover 81-92\% of the tactics they generate.

The current study is meant to be descriptive, rather than prescriptive; when we describe LLM responses as being templated, it is a factual description with no value judgment being passed. We also do not “endorse” the specific template we found—our study was not designed to find the most effective template, which will likely vary by many factors like context and culture—but instead sought to describe the modal template used by many of these commonly-used LLMs. But as mentioned, LLM responses are indeed well-liked by people, suggesting that the LLM template is perceived as very empathic. Perhaps over the course of their training, LLMs have indeed captured an “effective” set of empathic behaviors. One can also imagine that the average person could have a lot to learn from LLMs about how to be more empathic. A randomized experiment finds that peer supporters given access to an ‘editor’ AI that suggested changes produced messages that were more empathic and preferred to those in a control group without access to AI \citep{sharma2023human}, while a second study finds that practicing with an LLM-powered “coach” offering personalized feedback on empathy significantly improves participants’ communication \citep{kumar2026practicing}. Work in other domains like conflict resolution \citep{louie2024roleplay} have shown how LLMs have learnt effective patterns of human communication such that they can effectively roleplay and provide feedback---similarly, it is very plausible that these LLMs could be used as learning tools or coaches \citep{hecht2025using} to improve people’s empathic communication (e.g., during stressful work situations; \cite{das2025ai}). 

The current study also focuses on one type of templatedness, which is at the discourse functional level. That is, our tactics taxonomy defines a set of discourse functions that serve a role in structuring communication to achieve some goals or functional outcomes, which in our case is empathic support. This level of analysis complements previous work that show that LLM outputs are templated at the syntactic level \citep{shaib2024detection}, at the lexical level \citep{jiang2025artificial}, and even at a discourse structural level \citep{namuduri2025qudsim}. This is likely a function of how machine learning models compress information and learn to reproduce patterns in the data. That said, given that so many people are interacting with and using LLMs, and that LLM-generated text is appearing in more places that could influence future training of both LLMs and humans, it is concerning how this may result in greater homogenization of language. For instance, people already like AI-generated empathy \citep{ong2026ai, howcroft2025ai}; would people become used to this template---would they start to produce it more, and would they start to expect it more, both from their human communication partners as well as their AI models? If people start expecting other people to provide empathy in this manner, this would add another source of pressure to conform to these empathy templates, and perhaps result in a future where the human diversity in expressing empathy is replaced by these patterns \citep{sourati2026homogenizing}.

Our results also should not be interpreted to suggest that the LLM responses are the same regardless of contexts, like an automated voicemail message. Indeed, modern LLMs respond coherently, often referring back to what the empathy seeker has said (i.e., Paraphrasing). It is perhaps more that these responses are ``clich\'ed" (a lay, non-scientific term); LLM empathic responses follow this predictable cadence, which may not be immediately apparent the first few times one sees it, but will become noticeable after multiple interactions. That said, real-world LLM usage is also more complex than what we are able to study in these experiments. We are not able to study the effect of persona prompting, or memory effects (e.g., LLMs storing memory from previous interactions), on empathic responses. Presumably, if users regularly confide in their LLMs for social support, the LLM might learn their users' preferences and adjust its tactic usage accordingly. What we have described is perhaps just the baseline propensity for these LLMs when asked for empathy, which serves as a starting point for user interactions.

There are also downsides to templatedness. In our study we tried to diversify various contexts in which people seek empathy (various emotional situations encompassing workplace vs. relationships vs. other types of challenges). Consistently responding with a template across various contexts, almost by definition, suggests some degree of context insensitivity. If a model responds in the same manner regardless of the type of situation, then by definition it is not taking into account some features of the situation, and hence in some aspects it is not adapting its response fully to the situation. This could lead to undesired behaviors like AI sycophancy, which is when AI chatbots affirm their users even at the expense of factual accuracy \citep{sharma2024towards} or social consensus \citep{cheng2026elephant}. Indeed, recent research suggests that training LLMs to be warmer also causes them to be more sycophantic \citep{ibrahim2026training}, and that LLMs have difficulty distinguishing between users’ beliefs and facts \citep{suzgun2025language}, which provide evidence for this link between empathy and sycophancy. Moreover, recent research suggests that sycophancy may also be tied to maladaptive outcomes: Interacting with sycophantic chatbots may lead to increased attitude extremity and certainty \citep{rathje2025sycophantic}, decreased prosocial intentions \citep{cheng2026sycophantic}, and worsening mental health \citep{moore2025expressing, moore2026characterizing}. What makes this an especially tricky behavior to train out of models is that people prefer sycophantic to non-sycophantic models \citep{ong2026friendlier, rathje2025sycophantic, sharma2024towards}. These sycophantic interactions look a lot like empathic interactions—the chatbot incessantly validates and affirms the user—and what separates harmful sycophancy from beneficial empathy is context, and our findings on the templatic nature of LLM-generated empathy suggests one explanation as to why LLMs also produce sycophantic behavior, that in the extreme could lead to delusions and AI-induced psychosis \citep{moore2026characterizing}.

There are some limitations to our study. In developing our taxonomy, we defined several categories that we eventually discarded from our taxonomy because they appeared very infrequently. For instance, human writers sometimes expressed gratitude (“Thank you for sharing”; “I appreciate you trusting me”), used terms of endearment (“babe”, “girl”, “honey”), and spiritual references (“Praying for you”). These are ways of expressing empathy that may be specific to certain groups of people or may be a stylistic choice for the writer. Unfortunately, they appeared very infrequently even among our human data, suggesting perhaps that these tactics may be more idiosyncratic, so we decided to remove them from our final taxonomy. Due to this, and our limited sample size of human writing, our taxonomy may not be complete—for instance it may also miss out culturally-variable ways of expressing empathy, as the human writers in our first study were mainly from North America and Europe. That said, our final taxonomy achieves good coverage over our data: less than 3\% of the text in Study 1, and less than 1\% in Study 2, were rated as having no tactic from our taxonomy.


Another limitation is that we did not put in place strict measures to prevent our human writers’ use of AI in Study 1. We received very few responses that had obvious tells (changes in formatting that indicated a copied chunk of text) and we discarded those, but we adopted a lenient criterion when excluding such responses. We cannot rule out that some small portion of our human responses in Study 1 were generated by AI. Indeed, our analyses of template coverage also found that the “LLM template” matched 40\% of responses, and for those matches, covered 82\% of the responses generated. One interpretation of this finding is that there are people that naturally do use this template—or perhaps they may have been trained due to their psychology training—and so obviously we would expect some fraction of people to produce language that “looks like AI”. (Much like how some authors of this paper have been using em-dashes long before it became associated with AI writing). An alternative explanation is that some subset of these individuals may have been using LLMs—if this were true, and assume we were somehow able to identify and remove these LLM responses, the template matching results for the human writers would decrease further. This would only further strengthen our takeaway that human-written empathy is far more diverse compared to LLM-generated empathy, and does not affect our broader conclusions.

Finally, our research raises questions that should be explored by future research, especially on leveraging AI for well-being and mental health \citep{karnaze2026six}. Future research should explore how LLM responses vary (or can be made to vary) across cultures and recipient demographics (e.g., \cite{malik2025llms}), what are some of the prescriptive or normative recommendations for “effective” empathy, and how such insights and AI could be used to improve human flourishing via AI directly providing emotional support, or by using AI to help people to become more empathic and support those around them \citep{sharma2023human, kumar2026practicing, hecht2025using}.




\newpage
\section{Methods}\label{sec:Methods}

In this paper, we propose and validate a novel taxonomy of strategies or tactics that an individual empathy giver might use to express empathy toward empathy seekers. In Study 1, we develop and validate the taxonomy, to characterize patterns of human-written and LLM-generated empathic responses. In Study 2, we prompt an automated classifier to scale up our annotations, and report how these empathic response patterns generalize to a much larger dataset.

\subsection{Study 1: \textbf{Developing a Novel Taxonomy of Tactics for Expressing Empathy} }

In Study 1, we introduce a taxonomy of empathic response “tactics” that emphasizes explicit categories designed to be easily identifiable in text. We started with a literature review to identify previously-studied behaviors that may contribute to empathy to seed our codebook. We then collected a dataset of human-written (27 writers; 290 total responses) and LLM-generated empathic responses (3 models: \texttt{GPT-4 Turbo}, Llama-3.1-70b, and QWEN-2.5; 101 responses each for a total of 303 LLM responses) to 101 Reddit forum posts. Three raters then performed iterative thematic analysis to refine the codebook categories and definitions.

\paragraph{Literature Review} To seed our codebook, we first began by conducting a literature review to initially map out different components and facets of empathy \citep{cuff2016empathy,zaki2012neuroscience, perry2013understanding} and related constructs like social support 
\citep{taylor2011social, cohen1985stress, cutrona1987provisions, maisel2008responsive, thomas2026new, uchino1996relationship}, active listening \citep{rogers1957active} and supportive communication \citep{macgeorge2011supportive}. We particularly focused on the specific and concrete behaviors through which an empathy giver may express empathy (e.g., offering to donate time or money). For instance, “active listening” would not be concrete enough for our purposes; as commonly defined, this consists of behaviors like asking questions and paraphrasing what the other person has said, which are at our desired level of analysis. Our initial taxonomy were seeded with strategies like validation, empowerment, advice, paraphrasing, assistance, and information.

\paragraph{Data collection design}

We collected a set of 101 narratives from Reddit that spanned 5 contexts: romantic relationship, family/friend, work/school, travel, and mental health. 

\paragraph{Participants}

To collect a sample of human-written empathic responses, we recruited a sample of 33 psychology-degree (or related field) holders from Upwork, in two waves. Our Upwork responders had to meet the following criteria: (1) must have a background in psychology or related fields with a minimum of a bachelor’s degree, (2) must be fluent in English, (3) must be at least 18 years old. Upwork allows respondents to set their own payment rates, and our respondents' rates ranged between \$18-30/hour. The task was set to take 1.5-2 hours and was also the approximate average completion time, and responders wrote between 10-17 responses. Due to formatting issues with responses submitted by 6 responders, our final sample of human-written data included $n$ = 27 responders.

Responders on Upwork were presented with story vignettes (i.e., Reddit posts) and instructed to write a response as if they were a close friend of the narrator, and were encouraged to “give the best empathic response [they] can give”. They were not given a target length (word count).

The average age of our final sample of Upworkers was 32.3 years, with ages ranging from 21 to 60 years. 11.1\% of them self-identified as male, with no participants self-identifying as a third gender. 48.1\% of responders held a bachelor’s degree, 44.4\% held a master’s degree, and 7.4\% held a doctoral degree. 55.6\% of the sample identified as white, 18.5\% identified as Asian, 3.7\% identified as Middle Eastern/Arab, 11.1\% identified as Black, and 11.1\% identified as 2 or more races. 7.4\% of the sample also identified as Hispanic. 

\paragraph{LLM-generated responses}

We also generated responses to the same posts using three LLMs: \texttt{GPT-4 Turbo}, \texttt{Llama-3.1-70b}, and \texttt{QWEN-2.5}. The specific prompt was:

\begin{quote}
“You are a peer supporter. Read the support seekers’ post and write an appropriate and empathic response. Limit your response minimum 100 words to maximum 150 words. Do not exceed 150 words.”
\end{quote}

\paragraph{Perceived empathy ratings of responses}

As a measure of face validity, we ran a brief study to compare perceptions of empathy expressed in all $n$ = 593 human- and model-written responses. We replicated the study design from \citet{lee2024large}, recruiting 500 participants (mean age = 39.80 (SD = 13.11), 60.84\% female, 1.20\% non-binary/did not disclose) on Prolific to rate 4-8 responses each. For each rating trial, participants were presented with the original post from a support seeker, followed by an empathic response written by one of the four writers, and were then asked to rate the appropriateness and empathy-level of the response on a 5-point Likert scale. Responses were randomized so participants were unaware of the author. Due to the odd number of total responses across writers, participants were shown a mix of responses with at least one LLM- and one human-authored response in their trial set. Each response received ratings from at least 4 participants. To analyze the ratings, we fit a mixed-effects linear model predicting perceived empathy with the response writer as a fixed effect, and random effects by raters and response. 


\begin{figure}[t]
    \centering
    \includegraphics[width=0.55\linewidth]{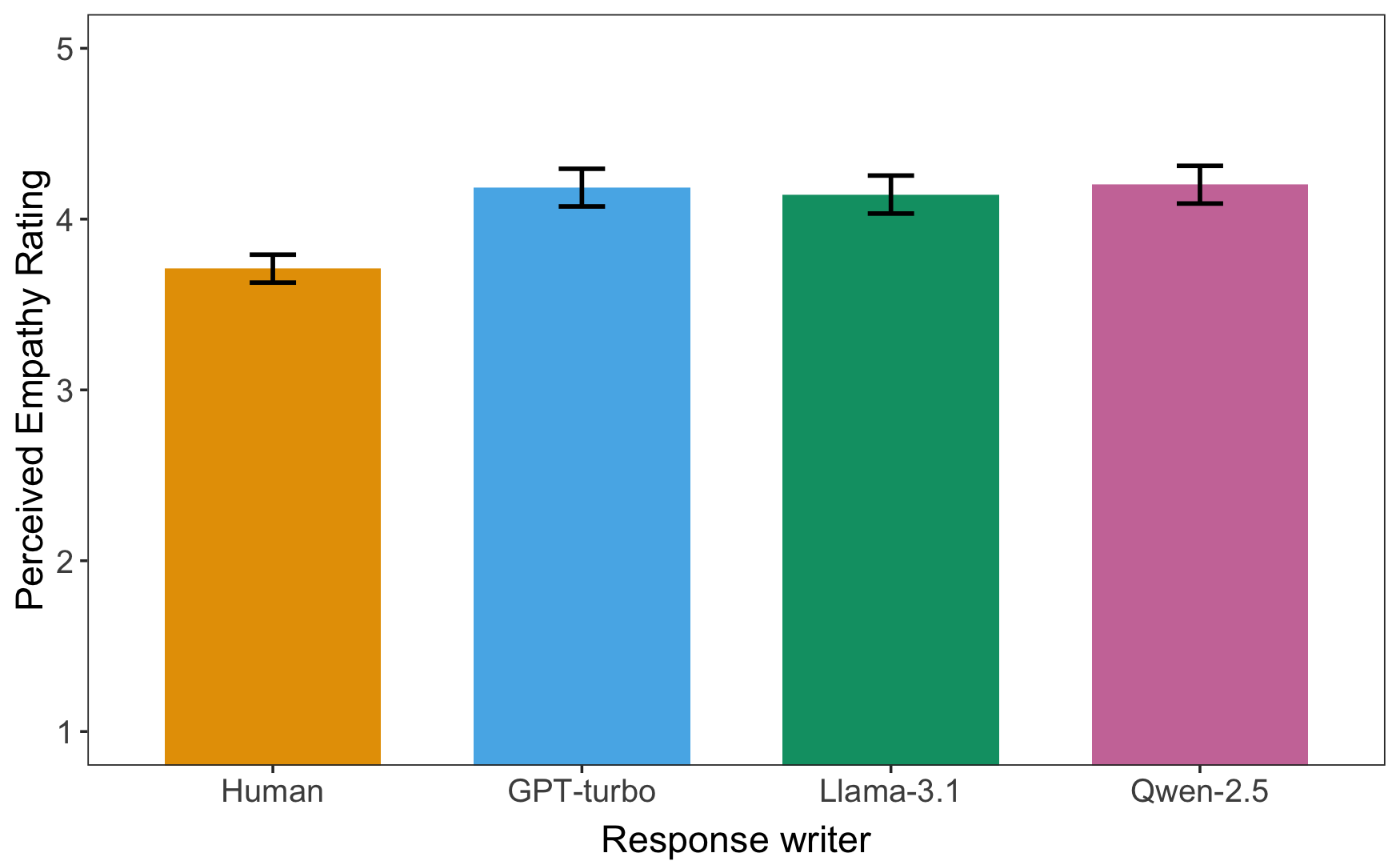}
    \caption{
    Study 1: Mean perceived empathy ratings by response writer with 95\% confidence intervals.
    }
    \label{fig:empathy}
\end{figure}

\paragraph{Coding Procedure}

We had three independent raters (two with bachelor’s degree and one undergraduate; one male) to code a sample of our responses on a sentence-by-sentence basis, for the presence of empathic tactics. Because it was difficult to determine when the same tactic appears consecutively, when each instance “begins” and “ends” (e.g., a sentence containing multiple pieces of advice), we only count one instance of that tactic by collapsing consecutive identical tactics into one (e.g., collapsing consecutive instances of Advice and counting it as just one instance of the Advice tactic). Raters also identified which part of the sentence was associated with which tactic, which gave us information about the order in which tactics appear. We did not allow tactics to overlap: each phrase could only be tagged with a maximum of one tactic.

The initial codebook was seeded with tactics that we identified from a literature search. Raters had regular meetings to discuss updates to the codebook, including adding or removing (e.g., combining) categories and refining the definitions and inclusion/exclusion criteria for the various categories. During this process, we noted several new tactics in the written responses, like Self-Disclosure. We then referred back to the literature to best integrate these strategies into our taxonomy.

After multiple rounds of our iterative qualitative coding process, we identified 10 empathic language tactics. The three raters then tagged 100 responses (with n=50 written by humans and n=50 written by GPT-4 Turbo, for a total of 787 sentences), and we calculated inter-rater reliability of tactics at the sentence level using Krippendorff’s alpha: The raters achieved an averaged IRR of $\alpha$=.78. Disagreements were resolved via consensus. The remainder of the data was then coded by one of the raters. 
We also supplemented our reliability analysis with a sequence-level agreement measure to capture similarity between coders in tactic ordering. We represented each coder's annotations as a collapsed sequence of tactics, just as we did for our main analyses, removing adjacent repetitions of the same label. We then calculated pairwise Levenshtein edit distance between coder sequences, where distance reflects the minimum number of insertions, deletions, or substitutions needed to align two sequences. We found that mean pairwise edit distance was 3.68 edits across collapsed sequences averaging 10.6 tactic units in length. 

\paragraph{Tactics taxonomy}

Our final taxonomy of 10 tactics, summarized in Table \ref{tab:taxonomy}, is presented in narrative form below, with the full codebook in the Supplemental Materials. The tactics were derived from iterative coding process described above rather than starting from predefined definitions of empathy. To further establish construct validity, we added a post-hoc mapping of the tactics to three commonly accepted facets of empathy: experience sharing (sometimes “affective empathy”), perspective taking (sometimes “cognitive empathy”), and empathic concern (sometimes “motivational empathy”). Under experience sharing, we have three tactics: producing an emotional expression, empowerment, and validation. Perspective-taking plays a significant role in four tactics: informational support-giving, paraphrasing what the empathy seeker said, offering reappraisal, and self-disclosing relevant information that the empathy giver has experienced. And empathic concern leads one to help, by offering advice or assistance, or by asking more questions to improve the empathy giver’s understanding of the other’s situation.

\textit{Emotional Expression.} An empathy-giver’s communication of their own feelings, reactions, or thoughts to the empathy-seeker as a result of hearing the empathy-seeker’s story. Expressing emotions like concern or compassion toward someone seeking support is an important way to show them that they (and their feelings) are being invested in. This is an integral part of building rapport and responding empathically \cite{elliott2011empathy}. Any use of emojis or emoticons in text is also considered an instance of this tactic.

\textit{Empowerment.} Positive, uplifting statements about the empathy-seeker’s character and capability to handle their given situation. Empowering an empathy-seeker through things like compliments can increase feelings of belonging and create a bond between them and the person they are speaking to \cite{zhao2021insufficiently}.

\textit{Validation.} Statements that reassure, normalize, or validate an empathy-seeker’s feelings. Research shows that validating someone’s feelings results in positive affect, particularly regarding the validation of physical pain \citep{edmond2015validating, linton2012painfully}. Validation produces positive affect and aids in establishing rapport between an empathy-giver and empathy-seeker.

\textit{Information.} Offering official resources that an empathy-seeker could turn to for help (e.g., links to websites, phone numbers, organizations), or stating information that may assist in answering the empathy-seeker’s questions, calming their anxieties, and potentially guiding them to a solution for their situation (if applicable).

\textit{Paraphrasing.} An empathy-giver’s perceived understanding of the situation, feelings, or experiences they inferred from the empathy-seeker. Particularly, we define an expression of Paraphrasing as an empathy-giver’s communication of the empathy-seeker’s feelings back to them. This is particularly important because an empathy-giver’s communication of their cognitive understanding establishes their invested interest in the empathy-seeker and serves as an expression of active listening, which is vital for forging trust and bonds between the two \citep{watson2007facilitating, glenn2024so}.

\textit{Reappraisal.} Statements that encourage the listener to reinterpret, reframe, or rethink their situation in a way that changes its emotional impact. An expression of reappraisal often introduces a new perspective that was not explicitly stated by the listener. We also consider general optimistic reframing about the future (e.g., ``everything will be okay”, ``life goes on") under our definition, similar to the reappraisal coding taxonomy in \citet{mcrae2012unpacking}.

\textit{Self-Disclosure.} This tactic refers to an empathy-giver sharing personal information about themselves or acknowledging similar past feelings and/or experiences to the empathy-seeker. Self-Disclosure is an integral component of relationship development and has been positively associated with relationship quality and satisfaction \citep{sprecher2004self}. Revealing personal information about oneself to another establishes intimacy, promotes openness, and fosters depth within that relationship. Additionally, it has been found that self-disclosure in online contexts is as effective as face-to-face contexts for relationship development \citep{dindia2011online}.

\textit{Advice.} Providing ideas for actionable solutions or coping strategies that the empathy-seeker could employ in the face of their situation. Giving advice has been linked to positive outcomes for the advice-giver \citep{eskreis2018dear}. Advice-giving has also been suggested to be an important part of being empathetic \citep{elliott2011empathy}.

\textit{Assistance.} Offering to personally do something for or with the empathy-seeker to aid them. This also includes offering personal contacts (friends/family/etc.) that could potentially aid the empathy-seeker. Research has found that helping results in positive consequences like feelings of belongingness and gratitude in those helped \citep{nadler1991help}. Essentially, assistance extends an invitation for help from the support-giver to the support-seeker.

\textit{Questioning.} Questions aimed at improving understanding of the empathy-seeker’s feelings, experiences, or situation. Asking questions for further clarification or more information indicates an active interest in the empathy-seeker, which is another important aspect of expressing empathy \citep{elliott2011empathy}.

\paragraph{Regular Expressions}

We used the formalism of regular expressions (“regex”) to classify the tactic templates in this study. Regular expressions are a way to describe a generic sequence of characters that are used to find patterns in text. For instance, \verb|[0-9]| defines a match set including all numerals from 0 to 9 and would match a single digit, and a 10-digit US phone number (ignoring dashes and spaces, and ignoring constraints on area codes) could be matched using \verb|[0-9]{10}|, meaning 10 instances of the match set \verb|[0-9]|. In this paper we represent empathic tactics using single letters (e.g., Paraphrasing as “P”, Validation as “V”), and represent a response as a sequence of characters denoting the sequence of tactics that the writer used: we have 10 tactics (in Table \ref{tab:taxonomy}), which are represented by 10 letters. 

We briefly introduce three relevant components of regular expressions that are used in this paper. A match set \verb|[PV]|, sometimes called character class, as introduced above, defines a match if any of the characters in the set are found (\textbf{P} or \textbf{V}). We can introduce quantifiers: \verb|+| and \verb|?|. The \verb|+| quantifier will match 1 or more of the preceding item. Thus, \verb|[PV]+| will match any string that contains \textbf{P} or \textbf{V} at least once, and would match sequences like: \textbf{V}, \textbf{P}, \textbf{VP}, \textbf{VPV}, \textbf{PVPV}, and so forth—importantly, it would not match the empty string, or strings containing other letters: On a sequence \textbf{PVPVAE}, it would only match the first four characters. The other quantifier \verb|?| will match 0 or 1 instance of the preceding item. So \verb|[XE]?| would match \textbf{X}, \textbf{E}, or the empty string. This allows a possible (but not mandatory) interjection of a single letter \textbf{X} or \textbf{E} at a particular position. Finally, \verb|^| denotes the start of the input, and \$ denotes the end of the input, so \verb|^P| will only match a \textbf{P} only if it is the first character in the input.

\textbf{Regular Expression Coverage.} We define two metrics to quantify the goodness of fit of a particular regular expression. We define the \textbf{coverage across responses} as the proportion of all responses which contained a sequence that was matched by a candidate regular expression. \textbf{Coverage within responses} was defined as the averaged proportion of each response accounted for by the candidate regular expression for responses that included a sequence matching the regular expression only. Thus, if a sequence of 10 characters had the first four characters matched by a regular expression, the within-response coverage for that response is 0.4; we calculate this value averaged across all responses where there is a match. Intuitively, shorter regular expressions maximize coverage across responses, while longer expressions maximize coverage within a response, so we attempt to balance maximizing both types of coverage (e.g., akin to maximizing precision and recall of a classification algorithm).

\textbf{Regular Expression Search.} We performed a search over the space of regular expressions using a mix of manual data inspection and a greedy search. We had several desiderata: for instance, we did not want to use wildcards \verb|(.*)| as that would trivially match everything, and we wanted to prioritize interpretability of the results. We used a greedy beam search to generate candidate template extensions by considering possible extensions that would maximize the harmonic mean of both across-response coverage and within-response coverage. For instance, on the initial search starting with the start-string character, the greedy search would evaluate candidates like \verb|^[X]|, \verb|^[P]|, \verb|^[E]|, ...  alongside possible quantifiers (\verb|?| and \verb|+| only, since \verb|*| would also trivially match everything) alongside their across-response and within-response coverage. We also considered groups of tactics that tended to co-occur together---for instance, \verb|P| and \verb|V| tended to co-occur, so we added the match set \verb|[PV]| into the space of possible candidates our search algorithm considered. We then inspected the top few candidates to decide on the final templates. 

In Table \ref{tab:regex} we report the across-response and within-response coverage for various patterns, which we numbered Pattern 1 through Pattern 5. As we constructed the patterns using an extension approach, the patterns are nested, i.e., Pattern 1 is contained within Pattern 2. Thus, a response that is matched by Pattern 2 will by definition be matched by Pattern 1 (with a lower within-coverage). But when we evaluate Pattern 2, we wanted to correctly count match responses that were \textbf{fully} matched by Pattern 1 (and which would not be matched by Pattern 2). Thus, after constructing Pattern 2, we formed a compound regex by concatenating Pattern 2 to the earlier patterns (Pattern 1) plus the ending character \verb|$|, to create the compound regular expression “\verb|<Pattern2>| $|$ \verb|<Pattern1>$|”, where $|$ is the ``or" operator. The row for Pattern 2 reports the matching results for this compound pattern. Similarly, for the row for Pattern 3, we created the compound expression “\verb|<Pattern3>| $|$ \verb|<Pattern2>$| $|$ \verb|<Pattern1>$|”, and so forth for Patterns 4 and 5.

\subsection{Study 2: Do “Templates” of Empathic Tactic Usage Generalize Across a Larger Sample?}


In our second study, we aimed to investigate whether these templates generalize across a larger dataset. We compared empathic responses written by humans (taken from Reddit), \texttt{GPT-4o}, \texttt{Llama3.3-70b-instruct}, and \texttt{Qwen3-32b}. The population of human responses differs from Study 1: Instead of using a small sample of participants with psychology backgrounds (and more responses per participant), we chose here to collect a more naturalistic sample of data: highly-rated Reddit comments.  The LLM models for Study 2 are updated versions (i.e., later generations) of the models from Study 1.

\paragraph{Data}

We collected a dataset of n = 1000 support seeking posts from Reddit. For each of these posts, we collected the most up-voted comment that was at least 100 words long, to serve as our human-written empathic response data. Posts and comments were scraped using Python’s \texttt{praw} package to connect with Reddit’s API and randomly selected based on the following criteria: (1) post must have been posted to one of seven target subreddits—r/work, r/confessions, r/emotionalsupport, r/offmychest, r/family, r/relationships, r/mentalhealth, (2) comment must contain at least 100 words, (3) comment must be the top-voted comment, (4) comment must have been posted between 2019 and 2024 . 

\paragraph{LLM-generated responses}

In addition to our human-written response data, we generated LLM-generated response data using \texttt{GPT-4o} (n = 962), \texttt{Llama3.3-70b-instruct} (n = 1000), and \texttt{Qwen3-32b} (n = 1000). When generating responses using GPT-4o, 38 stimuli were flagged for violating Microsoft Azure's API policy, resulting in a failure to generate responses to those Reddit posts.

\paragraph{Data Annotation Procedure}

Instead of using manual human coding as in Study 1, which was infeasible for this amount of data, we instead prompted a model to serve as an automated annotator. 
We prompted \texttt{claude-sonnet-4-5-20250929} as an automatic tagger of empathy tactics (i.e., llm-as-a-judge). We note that LLMs have been shown to be reliable at rating complex constructs, such as judging empathy according to various evaluative frameworks \citep{kumar2026large}; here we use an LLM to annotate the presence of specific empathic language behaviors. We chose to use a different model than any of the response-writing models, to avoid any possible bias. 

We utilized few-shot prompting with task-specific examples and in-context demonstrations. For a given input, the model was provided with the full list of taxonomy tactic definitions, annotation rules, in-context examples, and the whole empathic response. We deliberately simulated the same criteria and information that human annotators followed to ensure maximal consistency between human- and model-annotated tags. The model is instructed to decide whether each tactic is present in the given response and where it is present (which phrase/sentence/etc.), and was given the full empathic response to serve as both contextual background and the evaluation unit. It then produced labels for each section of the response it evaluated to best encompass an expression of a given tactic. See the full prompt in Supplemental Materials describing the definition and decision criteria for each tag.

\textit{Validation on Study 1 data}
To validate the tagger, we first ran the tagger on the human-annotated comparison dataset containing all 593 Study 1 responses, each annotated by phrase with the 10 empathy tactics. 
The automatic tagger achieved an average token-level F1 score of .70 across all the categories (advice: 0.84, assistance: 0.91, emotional expression: 0.57, empowerment: 0.63, information: 0.62, paraphrasing: 0.74, questioning: 0.81, reappraisal, 0.45: self disclosure: 0.83, validation, 0.59)
%


\paragraph{Data Analysis}

We used a similar data analysis procedure as in Study 1 to analyze the presence of tactics, and the sequences of tactics.

\backmatter

\section*{Ethics}
The studies conducted here were reviewed and approved by the Institutional Review Board at The University of Texas at Austin  (Protocol STUDY00004666). The source data were all publicly available and anonymous Reddit posts. 



\section*{Acknowledgements}

We thank Jiaying Liu and Katie Yan for their assistance on the project.

This material is based upon work supported by the National Science Foundation under Award No. 2443038 to D.C.O, and Awards 2107524 and 2145479 to J.J.L. Any opinions, findings and conclusions or recommendations expressed in this material are those of the author(s) and do not necessarily reflect the views of the National Science Foundation. 

The Toyota Research Institute partially supported this work. This article solely reflects the opinions and conclusions of its authors and not TRI or any other Toyota entity. 

This project has benefited from the Microsoft AI, Cognition, and the Economy (AICE) research program.

\begin{appendix}
\newpage

\section{Empathic Response Tactic Taxonomy Coding Guide}\label{sec:AppendixTactics}
Guidelines for coding empathic response tactics in text, in no particular order, with example statements for each tactic.

\subsection{Emotional Expression}
Statements where the speaker expresses their own feelings, reactions, opinions, or thoughts in response to the listener’s situation. \newline \newline
This includes:
\begin{enumerate}
    \item \textbf{Emotional responses} (e.g., \textit{“I’m so sorry”}, \textit{“I’m so happy for you”}, \textit{“That’s devastating”}, \textit{``That’s heartwarming to hear"})
    \item \textbf{Interjections} (e.g., \textit{“Wow”}, \textit{“Geez”})
    \item \textbf{Expressions of hope} (e.g., \textit{“I hope things get better for you”}, \textit{“I hope I can help”})
    \item \textbf{Personal reflections or opinions} (e.g., \textit{“I think you’ve put a lot of effort into this”}, \textit{“Your story made me think…”}, \textit{"I can't believe he did that"})
\end{enumerate} \underline{Important:} \newline \newline
Emotional Expression focuses on the speaker’s internal reaction, not on validating the listener’s feelings (that is Validation). \newline
Statements that ask the listener a question are NOT Emotional Expression (they are Questioning).
\begin{itemize}
    \item If a question implies a recommendation (e.g., \textit{“I wonder if you’ve considered…”}), it should be labeled as Advice (and possibly Questioning if both functions are present).
\end{itemize}
Statements that include actionable suggestions are NOT Emotional Expression (they are Advice). \newline
If the statement changes how the situation is interpreted (e.g., \textit{“things will improve”}), label as Reappraisal rather than Emotional Expression.

\subsection{Validation}
Statements that acknowledge, normalize, or affirm the listener’s feelings or experience. \newline \newline
This includes:

\begin{enumerate}
    \item \textbf{Explicit validation} (e.g., \textit{“I hear you”}, \textit{“That makes sense”})
    \item \textbf{Normalizing statements} (e.g., \textit{“This is normal”}, \textit{“People often feel this way”})
    \item \textbf{Generalizing difficulty} (e.g., \textit{“That’s hard”}, \textit{“College is hard”})
\end{enumerate} Validation can also be implicit, where the speaker frames the experience as understandable or common without explicitly saying \textit{“you are valid”}. \newline
Validation often appears as an evaluative frame before a paraphrased description of the listener's feelings or situation. In these cases, label the validating frame as Validation and the descriptive portion as Paraphrasing. \newline \newline
\underline{Important:} \newline \newline
\textit{“You’re not alone”} is Validation unless the speaker explicitly shares their own similar experience (That would be Self Disclosure). \newline
Statements that describe the listener’s situation or circumstances without affirming them are NOT Validation (those are Paraphrasing).

\subsection{Paraphrasing}
Statements where the speaker restates, summarizes, or interprets the listener’s situation, feelings, or experiences. \newline \newline
This includes:

\begin{enumerate}
    \item \textbf{Restating what the listener said} (e.g., \textit{“What I’m hearing is…”}, \textit{“It sounds like you…”}, \textit{"You said you just left your job"})
    \item \textbf{Interpreting the listener’s situation or emotions} (e.g., \textit{“You must be terrified”}, \textit{“It sounds like you did everything you could”})
    \item \textbf{Summarizing key elements of the listener’s story} (e.g., \textit{“It sounds like you and your roommate are having conflict”})
\end{enumerate}

\noindent Paraphrasing often follows Validation, Empowerment, or Emotional Expression within the same sentence. When another clause introduces or evaluates the listener's state, label only the descriptive portion as Paraphrasing. \newline \newline
\underline{Important:} \newline \newline
Paraphrasing focuses on reflecting or reconstructing the listener’s experience, not evaluating or normalizing it. \newline
Paraphrasing often mirrors the content of the listener’s message rather than introducing new information. \newline
Statements that primarily acknowledge or affirm feelings (e.g., \textit{“That’s really hard”},\textit{“That makes sense”}) are NOT Paraphrasing (they are Validation).
\begin{itemize}
    \item If the statement primarily describes the listener’s state, it is Paraphrasing
    \item If the statement primarily affirms or normalizes that state, it is Validation
\end{itemize}
Statements that express the speaker’s own emotions (e.g., \textit{“I’m so sorry”}) are NOT Paraphrasing (they are Emotional Expression). \newline
When Paraphrasing appears alongside other tactics, label only the portion that restates or interprets the listener’s experience.

\subsection{Advice}
Statements that provide actionable suggestions or recommendations for what the listener could do in response to their situation. \newline \newline
This includes:

\begin{enumerate}
    \item \textbf{Direct suggestions} (e.g., \textit{“You could try…”}, \textit{“It might help if you…”})
    \item \textbf{Hypothetical framing} (e.g., \textit{“If I were you…”}, \textit{“I would…”})
    \item \textbf{Lists or descriptions of possible actions} (e.g., \textit{“Things you could do include…”})
    \item \textbf{Suggestions involving mental actions} (e.g., \textit{“Try to think about it this way”})
\end{enumerate}

\noindent \underline{Important:} \newline \newline
Advice must include a clear or implied action for the listener to take. \newline
General statements without a suggested action are NOT Advice. \newline
Advice does NOT require the speaker to have personal experience \newline
If a sentence includes both personal experience and a recommendation, label them as separate spans:
\begin{itemize}
    \item The portion describing the speaker’s experience = Self Disclosure
    \item The portion providing a recommendation = Advice
\end{itemize}
If a sentence also functions as a question, split into separate spans when possible:
\begin{itemize}
    \item The suggestion portion = Advice
    \item The question portion = Questioning
\end{itemize}

\subsection{Assistance}
Statements where the speaker offers to personally help the listener or connect them to specific help. \newline \newline
This includes:

\begin{enumerate}
    \item \textbf{Direct offers of personal help} (e.g., \textit{“I can help”}, \textit{“Let me know if I can do anything”}, \textit{"We can figure this out"})
    \item \textbf{Invitations for continued support} (e.g., \textit{“I’m here for you”}, \textit{“You can talk to me anytime”})
    \item \textbf{Offers to take action on the listener’s behalf} (e.g., \textit{“I can look into this for you”})
    \item \textbf{Offering personal or specific contacts} (e.g., \textit{“My mom could pick you up”}, \textit{“I know someone who could help”})
\end{enumerate}

\noindent \underline{Important:} \newline \newline
Assistance requires an explicit or implied offer of the speaker’s own involvement or resources. \newline
General suggestions or recommendations (e.g., \textit{“You could try…”}) are NOT Assistance (they are Advice). \newline
Expressions of care without an offer of help (e.g., \textit{“I hope this helps”}, \textit{“I wish I could help”}) are NOT Assistance (they are Emotional Expression).

\subsection{Empowerment}
Statements that affirm the listener’s strength, worth, ability, or accomplishments. \newline \newline
This includes:

\begin{enumerate}
    \item \textbf{Affirming capability} (e.g., \textit{“You’ve got this”}, \textit{“You will figure it out”})
    \item \textbf{Recognizing effort or resilience} (e.g., \textit{“You did the best you could”}, \textit{“You’re doing great”})
    \item \textbf{Encouraging confidence in outcomes} (e.g., \textit{“You’ll get through this”}, \textit{“I’m sure you will succeed”})
    \item \textbf{Positive affirmations about the listener’s value} (e.g., \textit{“You deserve better”}, \textit{“They are lucky to have you”})
    \item \textbf{Celebratory or encouraging remarks} (e.g., \textit{“Congratulations!”}, \textit{“Good luck!”})
    \item \textbf{Expressions of pride} (e.g., \textit{“I’m proud of you”})
\end{enumerate}

\noindent Empowerment may appear as a short praising frame (e.g., \textit{"it's great that"}) before a longer paraphrased description of the listener's actions or insight. In these cases, split the praising frame as Empowerment and the descriptive clause as Paraphrasing. \newline \newline
\underline{Important:} \newline \newline
Statements expressing pride in the listener (e.g., \textit{“I’m proud of you”}) should always be labeled as Empowerment. \newline
Empowerment focuses on uplifting the listener’s ability, worth, or accomplishments. \newline
Statements that primarily describe the listener’s feelings are NOT Empowerment (they are Paraphrasing or Validation). \newline
Statements that express the speaker’s emotions without affirming the listener’s ability or worth are NOT Empowerment (they are Emotional Expression). \newline
Statements that provide specific actions or suggestions are NOT Empowerment (they are Advice). \newline
Do not label statements that primarily acknowledge difficulty (e.g., \textit{“that’s really hard”}) as Empowerment (those are Validation).

\subsection{Questioning}
Statements that ask the listener for information, clarification, or reflection about their feelings, experiences, or situation. \newline \newline
This includes:

\begin{enumerate}
    \item \textbf{Direct questions} (e.g., \textit{“What happened?”}, \textit{“How are you feeling?”}, \textit{“Are you alright?”})
    \item \textbf{Indirect or embedded questions} (e.g., \textit{“I wonder if…”}, \textit{“I would ask if…”})
    \item \textbf{Questions that prompt reflection} (e.g., \textit{“Do you think this will have value?”})
\end{enumerate}

\noindent \underline{Important:} \newline \newline
Questioning focuses on seeking information or prompting reflection from the listener. \newline
Questioning does NOT require a question mark; intent matters more than punctuation. \newline
If a statement contains both a question and a suggested action (e.g., \textit{“Have you tried talking to someone”}), split into separate spans when possible:
\begin{itemize}
    \item The question portion = Questioning (e.g., \textit{“Have you tried…”})
    \item The suggestion portion = Advice (e.g., \textit{“talking to someone”})
\end{itemize}

\subsection{Reappraisal}
Statements that encourage the listener to reinterpret, reframe, or rethink their situation in a way that changes its emotional impact. \newline \newline
This includes:

\begin{enumerate}
    \item \textbf{Reframing the meaning of the situation} (e.g., \textit{“It wasn’t your fault”}, \textit{“Some things are beyond our control”})
    \item \textbf{Shifting perspective} (e.g., \textit{“This is temporary”}, \textit{“There will be other opportunities”})
    \item \textbf{Encouraging alternative ways of thinking} (e.g., \textit{“Maybe you’ll think differently about this over time”})
    \item \textbf{Emphasizing change, growth, or future possibilities} (e.g., \textit{“This will pass”}, \textit{“There is still so much life ahead”})
\end{enumerate}

\noindent \underline{Important:} \newline \newline
Reappraisal involves changing how the situation is interpreted, not just expressing positivity. \newline
Reappraisal often introduces a new perspective that was not explicitly stated by the listener. \newline
Statements that only encourage or uplift (e.g., \textit{“You’ve got this”}) are NOT Reappraisal (they are Empowerment). \newline
Statements that normalize feelings without changing perspective (e.g., \textit{“that’s really hard”}) are NOT Reappraisal (they are Validation). \newline
Statements that suggest an action (e.g., \textit{“try not to think about the past”}) are Advice; any accompanying justification (e.g., \textit{“because it is now in the past”}) may be Reappraisal and should be labeled separately. \newline
General optimistic reframing about the future (e.g., \textit{“Everything will be okay”}) should be treated as Reappraisal.

\subsection{Self Disclosure}
Statements where the speaker shares personal experiences or background information about themselves. \newline \newline
This includes:

\begin{enumerate}
    \item \textbf{Directly stating similar experiences} (e.g., \textit{“I’ve felt the same way”}, \textit{“That happened to me too”})
    \item \textbf{Referencing personal situations} (e.g., \textit{“When I was in college…”}, \textit{“My daughter…”}, \textit{“I visited Thailand…”})
    \item \textbf{Describing how they personally respond to similar feelings} (e.g., \textit{“When I feel this way, I…”})
\end{enumerate}

\noindent \underline{Important:} \newline \newline
Self Disclosure requires explicit reference to the speaker’s own experience or life. \newline
General advice or suggestions (e.g., \textit{“In this case I would recommend…”}) are NOT Self Disclosure unless the speaker clearly ties the advice to their own experience.

\subsection{Information}
Statements that provide factual, general, or explanatory content relevant to the listener’s situation. \newline \newline
This includes:

\begin{enumerate}
    \item \textbf{General knowledge or explanations} (e.g., \textit{“Flying is the safest form of travel”}, \textit{“Migraines can be caused by stress”})
    \item \textbf{Statements about how things work or why something happens} (e.g., \textit{“Sunshine is good for your mental health”})
    \item \textbf{Cultural or situational facts} (e.g., \textit{“It’s customary to kiss on both cheeks in French culture”})
    \item \textbf{Providing external resources} (e.g., links, phone numbers, organizations)
\end{enumerate}

\noindent Information often follows Advice as a justification or explanation. \newline \newline
\underline{Important:} \newline \newline
Information does NOT include actionable suggestions (those are Advice). \newline
Information does NOT include personal experiences (those are Self Disclosure). \newline
Do not label statements that primarily acknowledge or normalize feelings as Information (those are Validation). \newline
When Advice and Information appear together, label them as separate spans:
\begin{itemize}
    \item The action = Advice
    \item The explanation or justification = Information
\end{itemize}

\vspace{1em}
\hrule
\vspace{1em} 

\noindent \underline{\textbf{Key tactic distinctions:}}
\begin{itemize}
    \item \textit{Validation} = affirming feelings
    \item \textit{Self Disclosure} = sharing personal experience
    \item \textit{Advice} = actionable recommendations
    \item \textit{Assistance} = offering involvement
    \item \textit{Information} = providing general facts or explanations
    \item \textit{Emotional Expression} = affective responses
    \item \textit{Empowerment} = positive encouragement
    \item \textit{Reappraisal} = changing interpretations
    \item \textit{Paraphrasing} = reflecting back what happened
    \item \textit{Questioning} = information seeking
\end{itemize}

\newpage

\section{Other Empathic Tactics}\label{sec:AppendixExtraTactics}
Descriptions of empathic tactics we observed in response data across both of our studies that were not included in final analyses due to low base rates of appearance.

\begin{table}[!ht]
\centering
\begin{tabular}{p{2.0cm}p{5.0cm}p{4.5cm}}
\hline
\textbf{Tactic}   & \textbf{Description}    & \textbf{Examples}  \\\hline
Gratitude        & \hangindent=1em Thanking or appreciating the empathy-seeker for sharing their story.   &  \hangindent=1em \emph{Thank you for sharing.} $|$ \emph{I appreicate you.} \\
Spirituality       & \hangindent=1em A religious statement of support to the empathy-seeker.   &  \hangindent=1em \emph{Praying for you.} $|$ \emph{In another life} $|$ \emph{Karma will get them.}\\
\hangindent=1em Terms of Endearment     & \hangindent=1em Nicknames expressing love, care, or affection toward the empathy-seeker.   &  \hangindent=1em \emph{Bro} $|$ \emph{Friend} $|$ \emph{Bestie} \\
Solidarity      & \hangindent=1em A statement from the empathy-giver that they are \textit{presently} going through the same situation as the empathy-seeker.  & \hangindent=1em \emph{We're in this together.} $|$ \emph{Same for me right now.}\\
\hline
\end{tabular}
\caption{List of tactics, their descriptions and example statements.}
\label{tab:taxonomy-extended}
\end{table}

\paragraph{Terms of Endearment.}
Nicknames expressing love, care, or affection toward the empathy-seeker. The usage of endearing terms typically indicates established closeness between two people. However, expressing a Term of Endearment generally promotes closeness and intimacy in relationships, and can occur even between consequential strangers within the context of online communities \citep{febrianti2021millennials}.  

Example statements of this tactic include:

\begin{itemize}
\item \emph{Bro, Friend, Love, Bestie}
\end{itemize}

\paragraph{Gratitude.}
Thanking or appreciating the empathy-seeker for sharing their story. Research has indicated that expressing gratitude facilitates and maintains relationships \citep{algoe2008beyond, bartlett2012gratitude}. The promotion of positive social relationships following an expression of Gratitude allows an empathy-giver to establish a bond with an empathy-seeker. 

Example statements of this tactic include:

\begin{itemize}
\item \emph{Thank you for trusting me.}
\item \emph{I appreciate you.}
\item \emph{I'm grateful you shared this with me.}
\end{itemize}

\paragraph {Spirituality.}
A religious statement of support to the empathy-seeker. Research has shown that religious expression is used as a means of support-giving, though there are mixed results about whether positive outcomes result from its use \citep{lieberman2009relationship}. However, independent of its potential impact on an empathy-seeker, the prevalence of religious expression in support statements necessitates Spirituality as a category in our framework. 

Example statements of this tactic include:

\begin{itemize}
\item \emph{I'm praying for you.}
\item \emph{God bless you.}
\item \emph{In your next life...}
\item \emph{Karma is real.}
\end{itemize}

\paragraph{Solidarity.}
A statement from the empathy-giver that they are \textit{presently} going through the same or a similar situation as the empathy-seeker. An expression of Solidarity provides reassurance to the support-seeker that they are not alone in their given situation. Like Self-Disclosure, an expression of Solidarity reveals personal information about the support-giver to the support-seeker, which is important for fostering trust and intimacy between them \citep{wheeless1977measurement}. We consider Solidarity to be a type of Self-Disclosure; it is a statement that reveals personal information about the empathy-giver specifically expressing that they are \textbf{presently} also in the same or a similar boat. Importantly, while there is always Self-Disclosure in an expression of Solidarity, there is not always Solidarity in an expression of Self-Disclosure. 

Example statements of this tactic include:

\begin{itemize}
\item \emph{We're in this together.}
\item \emph{I'm struggling with this right now too.} 
\item \emph{We'll suffer together.}
\end{itemize}

\newpage

\section{Study 1: Response Writer Prompts}

The following figures detail the prompts given to human and LLM response writers in Study 1.

\begin{figure}[h]
     \centering
     \begin{tcolorbox}[colback=gray!10!white,colframe=blue!50!black,
         colbacktitle=blue!40!black,
         title=\textbf{Prompt Instructions for Human Response Writers},
         width=0.95\textwidth
     ]
    
     \raggedright 

     \textbf{\texttt{\#\#\#} Instruction:}\smallskip

        ``You will have \textbf{*X} stories/issues of varying topics in an Excel workbook. Each story will be separated into individual sheets in this workbook. You will read each one and write out an empathic response to the narrator of the story in the B column.\newline \smallskip
    
        Write as if you are a close friend of the narrator. You may identify what the narrator is feeling first and give out the best empathetic response you can give. The length does not matter and do not be pressured to write a long response either."\smallskip

     \end{tcolorbox}
    
     \caption{Prompt given to human response writers recruited through Upwork in Study 1. \textbf{*X=16-17 in Wave 1 and X=10 in Wave 2.}}
     \label{fig:prompt-upwork}
 \end{figure}

During study 1, we collected empathic responses from humans in two waves. In the first wave, we recruited 3 human respondents through Upwork to write empathic messages to the narrators of 50 posts taken from several Reddit forums spanning across our five main contexts of interest. In the second wave, we recruited 30 more respondents to write empathic messages to the narrators of 95 posts from Reddit. For generalizability and relevance, we removed the “travel” context (n = 6 posts) from our initial set of 50 posts and added an additional 46 posts to our stimuli set, totaling to n = 95 unique narratives that we presented to our second wave of response writers. Figure \ref{fig:prompt-upwork} shows the exact protocol given to recruited Upworkers in both waves.

\begin{figure}[h]
     \centering
     \begin{tcolorbox}[colback=gray!10!white,colframe=blue!50!black,
         colbacktitle=blue!40!black,
         title=\textbf{Prompt Instructions for LLM Response Writers},
         width=0.95\textwidth
     ]
    
     \raggedright 

     \textbf{\texttt{\#\#\#} Instruction:}\smallskip

        “You are a peer supporter. Read the support seekers’ post and write an appropriate and empathic response. Limit your response minimum 100 words to maximum 150 words. Do not exceed 150 words.”\smallskip

     \end{tcolorbox}
    
     \caption{Prompt given to GPT4-turbo, QWEN-2.5, and Llama3.1-70b in Study 1.}
     \label{fig:prompt-llm}
 \end{figure}

 In study 1, we also generated empathic responses using 3 LLMs. Figure \ref{fig:prompt-llm} shows the prompt given to LLM writers.

\newpage
\section{Study 2: Prompting an Automated Tagger}

The prompt for LLM Qualitative Tagger used in Study 2 (\texttt{claude-sonnet-4-5-20250929}) consisted of four parts that are appended together into a single prompt. The actual prompt did not have these parts demarcated, but we describe them here separately for conceptual clarity.

\begin{verbatim}
[Intro]
[Tactic Definitions]
[Additional Instructions]
[Few Shot Examples]
\end{verbatim}

\subsection{[Intro]}

The [Intro] chunk consists of the following two sentences:

\begin{quotation}
You are a careful qualitative researcher. Your task is to annotate the comment by identifying spans of text that match the tactic definitions below.
\end{quotation}

\subsection{[Tactic Definitions]}

The tactic definitions chunk is Supplemental Materials \ref{sec:AppendixTactics}.


\subsection{[Additional Instructions]}

\begin{center}
\begin{tcolorbox}[
    enhanced,
    breakable,
    colback=gray!10!white,
    colframe=red!50!black,
    colbacktitle=red!40!black,
    title=\textbf{[Additional Instructions] part of Prompt for LLM Qualitative Tagger},
    width=\textwidth
]

\footnotesize
\raggedright
            
         \medskip
    
        CORE RULE: Annotate functional spans, not whole sentences.
        
        \begin{itemize}[noitemsep, topsep=2pt]
            \item A sentence may contain multiple tactics.
            \item If different parts of a sentence serve different functions, you MUST split them.
            \item Do NOT assign a single label to a full sentence if multiple functions are present.
        \end{itemize}
        
        A span must:
        \begin{itemize}[noitemsep, topsep=2pt]
            \item be a phrase or clause within a single sentence
            \item appear EXACTLY in the original text
            \item be the smallest meaningful unit expressing that tactic
        \end{itemize}
        
        \medskip
        KEY PATTERN (VERY IMPORTANT)
        \medskip
        
        Many sentences follow this structure:
        \texttt{[short interpersonal frame] + [description of the listener's situation]}
        
        You MUST split these.
        
        Rules:
        \begin{itemize}[noitemsep, topsep=2pt]
            \item The first part (evaluating, reacting, or praising) = label as Validation, Empowerment, or EmotionalExpression
            \item The second part (describing the listener’s situation) = label as Paraphrasing
        \end{itemize}
        
        Examples of frames:
        \begin{enumerate}[noitemsep, topsep=2pt]
            \item "it's understandable that" = Validation
            \item "it's okay to" = Validation
            \item "I believe you" = Validation
            \item "it's great that" = Empowerment
            \item "I'm sorry that" = EmotionalExpression
            \item "that must feel" = Validation
        \end{enumerate}
        
        \medskip
        KEY DISTINCTION:
        \medskip
        
        Validation = affirms or evaluates the listener’s feelings\newline
        Paraphrasing = describes or restates the listener’s situation\newline
        
        If both occur in the same sentence, you MUST split:
        \begin{itemize}[noitemsep, topsep=2pt]
            \item evaluative part = Validation
            \item descriptive part = Paraphrasing
        \end{itemize}
        
        \medskip
        IMPORTANT: Do NOT overuse Empowerment
        \smallskip
        
        Empowerment ONLY applies when: explicitly praising or affirming strength, progress, or ability
        
        NOT Empowerment:
        \begin{itemize}[noitemsep, topsep=2pt]
            \item "it's okay to..." = Validation
            \item "it's understandable that..." = Validation
            \item "I believe you" = Validation
        \end{itemize}
        
        If unsure, prefer Validation over Empowerment.
        
        \medskip
        MANDATORY CHECK:
        \medskip
        
        For each sentence:
        Does this contain more than one function?
        
        If yes:
        = split into multiple spans
        
        \medskip
        ANNOTATION RULES:
        \medskip
        
        \begin{enumerate}[noitemsep, topsep=2pt]
            \item Identify all spans matching the categories
            \item Preserve order
            \item Do NOT paraphrase text
            \item If multiple tactics apply, choose the most specific
        \end{enumerate}

        \medskip
        Return JSON in exactly this format:
        
        \begin{verbatim}
        {"segments": [
            {
              "order": 1,
              "text_span": "exact text from comment",
              "label": "CATEGORY_NAME"
            }
          ]
        }
        \end{verbatim}
                
\end{tcolorbox}

\smallskip
\textbf{Figure 5.} Additional Instructions part of prompt given to qualitative tagger
\texttt{claude-sonnet-4-5-20250929} in Study 2.
\label{fig:prompt-tagger}
\end{center}

\subsection{[Few Shot Examples]}\label{sec:AppendixFewShot}

\begin{center}
\begin{tcolorbox}[
    enhanced,
    breakable,
    colback=gray!10!white,
    colframe=purple!50!black,
    colbacktitle=purple!40!black,
    title=\textbf{[Few Shot Examples] part of Prompt for LLM Qualitative Tagger},
    width=\textwidth
]

\footnotesize
\raggedright

     \textbf{\texttt{\#\#\#} Few Shot Examples:}\smallskip

    \begin{verbatim}
        Example 1
        
        Comment:
        "I’m really sorry that you’ve been feeling stuck in something you don’t enjoy."
        
        Output:
        {
          "segments": [
            {
              "order": 1,
              "text_span": "I’m really sorry that",
              "label": "EmotionalExpression"
            },
            {
              "order": 2,
              "text_span": "you’ve been feeling stuck in something you don’t enjoy.",
              "label": "Paraphrasing"
            }
          ]
        }
        
        ---
        
        Example 2
        
        Comment:
        "It's really impressive that you kept working on this for so long."
        
        Output:
        {
          "segments": [
            {
              "order": 1,
              "text_span": "It's really impressive that",
              "label": "Empowerment"
            },
            {
              "order": 2,
              "text_span": "you kept working on this for so long.",
              "label": "Paraphrasing"
            }
          ]
        }
        
        ---
        
        Example 3
        
        Comment:
        "It's understandable that you're feeling overwhelmed right now."
        
        Output:
        {
          "segments": [
            {
              "order": 1,
              "text_span": "It's understandable that",
              "label": "Validation"
            },
            {
              "order": 2,
              "text_span": "you're feeling overwhelmed right now.",
              "label": "Paraphrasing"
            }
          ]
        }
        
        ---
        
        Example 4
        
        Comment:
        "I believe you, and it sounds like you've been dealing with this for a long time."
        
        Output:
        {
          "segments": [
            {
              "order": 1,
              "text_span": "I believe you,",
              "label": "Validation"
            },
            {
              "order": 2,
              "text_span": "and it sounds like you've been dealing with this for a long time.",
              "label": "Paraphrasing"
            }
          ]
        }
        
        ---
        
        Example 5
        
        Comment:
        "That must feel really frustrating, and I’m sorry you’re going through that."
        
        Output:
        {
          "segments": [
            {
              "order": 1,
              "text_span": "That must feel really frustrating,",
              "label": "Validation"
            },
            {
              "order": 2,
              "text_span": "and I’m sorry",
              "label": "EmotionalExpression"
            },
            {
              "order": 3,
              "text_span": "you’re going through that.",
              "label": "Paraphrasing"
            }
          ]
        }
        
        ---
        
        Example 6
        
        Comment:
        "It's okay to want time for yourself."
        
        Output:
        {
          "segments": [
            {
              "order": 1,
              "text_span": "It's okay to want time for yourself.",
              "label": "Validation"
            }
          ]
        }
        
        ---
        
        Example 7
        
        Comment:
        "You’ve shown a lot of strength in handling this."
        
        Output:
        {
          "segments": [
            {
              "order": 1,
              "text_span": "You’ve shown a lot of strength in handling this.",
              "label": "Empowerment"
            }
          ]
        } 
        
        ---
        
        Incorrect Example 1 (DO NOT DO THIS)
        
        Comment:
        "It's understandable that you're feeling overwhelmed."
        
        Incorrect Output:
        {
          "segments": [
            {
              "order": 1,
              "text_span": "It's understandable that you're feeling overwhelmed.",
              "label": "Validation"
            }
          ]
        }
        
        Correct Output:
        {
          "segments": [
            {
              "order": 1,
              "text_span": "It's understandable that",
              "label": "Validation"
            },
            {
              "order": 2,
              "text_span": "you're feeling overwhelmed.",
              "label": "Paraphrasing"
            }
          ]
        }

    \end{verbatim}

\end{tcolorbox}

\smallskip
\textbf{Figure 6.} Few-shot examples in prompt given to qualitative tagger
\texttt{claude-sonnet-4-5-20250929}.
\label{fig:few-shot-tagger}
\end{center}
     
\newpage

\end{appendix}





\bibliography{references}


\end{document}